\documentclass{article}

\usepackage[preprint]{tagds_2025}
\usepackage{graphicx}
\usepackage{subcaption}
\usepackage{comment}
\usepackage{enumitem}
\usepackage{wrapfig}
\usepackage{float}

 \workshoptitle{Topology, Algebra, and Geometry in Data Science}

\usepackage[utf8]{inputenc} 
\usepackage[T1]{fontenc}    
\usepackage{hyperref}       
\usepackage{url}            
\usepackage{booktabs}       
\usepackage{amsfonts}       
\usepackage{nicefrac}       
\usepackage{microtype}      
\usepackage{xcolor}         

\newcommand{\XC}[1]{{{\textcolor{purple}{#1}}}}

\title{Neural Local Wasserstein Regression}

\author{%
  Inga Girshfeld \\
  Department of Mathematics\\
  University of Southern California\\
  Los Angeles, CA 90007 \\
  \texttt{girshfel@usc.edu} \\ \And Xiaohui Chen \\
  Department of Mathematics\\
  University of Southern California\\
  Los Angeles, CA 90007 \\
  \texttt{xiaohuic@usc.edu} \\
}

\begin{document}

\maketitle

\begin{abstract}
    We study the estimation problem of distribution-on-distribution regression, where both predictors and responses are probability measures. Existing approaches typically rely on a global optimal transport map or tangent-space linearization, which can be restrictive in approximation capacity and distort geometry in multivariate underlying domains. In this paper, we propose the \emph{Neural Local Wasserstein Regression}, a flexible nonparametric framework that models regression through locally defined transport maps in Wasserstein space. Our method builds on the analogy with classical kernel regression: kernel weights based on the 2-Wasserstein distance localize estimators around reference measures, while neural networks parameterize transport operators that adapt flexibly to complex data geometries. This localized perspective broadens the class of admissible transformations and avoids the limitations of global map assumptions and linearization structures. We develop a practical training procedure using DeepSets-style architectures and Sinkhorn-approximated losses, combined with a greedy reference selection strategy for scalability. Through synthetic experiments on Gaussian and mixture models, as well as distributional prediction tasks on MNIST, we demonstrate that our approach effectively captures nonlinear and high-dimensional distributional relationships that elude existing methods.
\end{abstract}

\section{Introduction}

Modeling relationships where both predictors and responses are probability measures is an emerging challenge in statistics and machine learning. Such \emph{distribution-on-distribution} (DoD) regression arises naturally in diverse applied domains where data are inherently measure-valued. In computer vision, images can be processed as two-dimensional histograms of grayscale pixel levels (or in the RGB space for colored images), and the task of super-resolution is to constructing a high-resolution image as the response distribution from a given low-resolution image as the predictor \citep{kim2010single,lai2017deep}. In biomedical sciences, population heterogeneity such as mortality rate is often captured through empirical age-at-death distributions \citep{Chiou01062009,Shang03042017}. In fluid dynamics, researchers routinely compare and predict distributions of spatiotemporal fields such as temperature and precipitation in climate sciences \citep{10.5555/3433701.3433712,li2021fourierneuraloperatorparametric}. Across these settings, regression directly in the space of measures provides a principled framework for capturing complex distributional relationships that methods designed for Euclidean data cannot accommodate.

A central difficulty is that probability measures reside in a non-Euclidean space. A large body of recent work has sought to endow such spaces with meaningful geometry through the theory of optimal transport (OT) \citep{villani2009ot,santambrogio2015ot}. One line of work defines regression through \emph{global} OT maps. \citep{ghodrati2022distribution,ghodrati2022minimax,ghodrati2023transportation} formalizes DoD regression by transporting the predictor distribution via a Monge map, with minimax analysis and Gaussian extensions. \citep{okano2024distribution} further specializes this framework to Gaussian families via the Bures–Wasserstein geometry. An alternative direction leverages \emph{tangent-space linearization}, notably Wasserstein regression \citep{chen2023wasserstein}, which lifts measures to tangent bundles, performs regression in linearized coordinates, and maps predictions back. More broadly, the Fr\'echet regression framework \citep{petersen2019frechet} provides kernel-based estimators for random objects, with inference tools such as Wasserstein F-tests on Bures–Wasserstein manifolds \citep{xu2024wasserstein}. A third strand concerns \emph{conditional optimal transport}, where maps depend on covariates: data-driven conditional OT \citep{tabak2021data}, conditional Brenier's map via entropic regularization \citep{baptista2024conditional}, and neural conditional maps \citep{wang2023neuralcot}.

While these approaches provide important foundations, they share key limitations. Global-map methods \citep{pmlr-v28-oliva13,ghodrati2022distribution,ghodrati2022minimax,ghodrati2023transportation} impose strong structural assumptions and are difficult to scale beyond one-dimensional or Gaussian data. Tangent-space methods \citep{chen2023wasserstein,petersen2019frechet} depend on linearization at the Fr{\'e}chet means of predictor and response measures, which can distort geometry in higher dimensions than the univariate case. Conditional OT \citep{tabak2021data,baptista2024conditional,wang2023neuralcot} defines maps indexed by covariates but still assumes each map is globally defined. To the best of our knowledge, no prior DoD regression framework explicitly models the transport as \emph{locally defined} over subsets of the source measure’s support.

In this paper, we introduce a new framework for nonparametric Wasserstein regression where both source and target are probability measures. Our model departs from the prevailing paradigms by allowing the transport between measures to be only \emph{locally} defined, rather than enforcing a single global map or a linearized approximation. This local perspective not only broadens the class of admissible transformations (cf. Section~\ref{subsec:regression_model} for the counterexample of the global map) but also offers robustness in higher-dimensional settings where global maps may fail to exist or be statistically unstable.

\subsection{Our Contribution}

We introduce a new framework for modeling nonparametric Wasserstein regression where both source and target are probability measures. We propose the \emph{neural local Wasserstein regression}, a statistical framework for DoD regression based on locally defined transport maps that generalizes beyond one-dimensional or Gaussian data and avoids restrictive global-map assumptions and linearization structures. We demonstrate the flexibility of our method on synthetic and real examples and illustrate how local-map regression can capture complex higher-dimensional distributional relationships.

\section{Background}

\subsection{Wasserstein Distance and Optimal Transport}

Let $\mathcal{P}_2(\mathbb{R}^d)$ denote the space of Borel probability measures on $\mathbb{R}^d$ with finite second moment. The squared \emph{2-Wasserstein distance} between $\mu, \nu \in \mathcal{P}_2(\mathbb{R}^d)$ is defined as
\[
W_2^2(\mu,\nu) = \inf_{\pi \in \Pi(\mu,\nu)} \int_{\mathbb{R}^d \times \mathbb{R}^d} \|x-y\|^2 \, d\pi(x,y),
\]
where $\Pi(\mu,\nu)$ is the set of couplings with marginals $\mu$ and $\nu$. This defines a metric on $\mathcal{P}_2(\mathbb{R}^d)$ that encodes both distributional and geometric information \citep{villani2009ot,santambrogio2015ot}. When $\mu$ is absolutely continuous with respect to Lebesgue measure, the \emph{Brenier theorem} guarantees that there exists a unique optimal transport map $T: \mathbb{R}^d \to \mathbb{R}^d$ such that $T_{\sharp}\mu=\nu$ and
\[
W_2^2(\mu,\nu) = \int_{\mathbb{R}^d} \|x - T(x)\|^2 \, d\mu(x),
\]
and this map is characterized as the gradient of a convex potential \citep{brenier1991polar}. These properties make the Wasserstein space a Riemannian-like metric space, with geodesics given by displacement interpolation \citep{ambrosio2008gradient,panaretos2020invitation}. The Wasserstein distance has become a fundamental tool in statistics and machine learning, providing a principled way to compare and model distributions in fields ranging from generative modeling to functional data analysis \citep{peyre2019computational,panaretos2020invitation}. In the context of regression, it allows us to directly model relationships between probability measures as geometric objects.

\subsection{Classical Nonparametric Regression}

In Euclidean settings, regression between random variables $Y \in \mathbb{R}$ and predictors $X \in \mathbb{R}^d$ has been extensively studied. A cornerstone is \emph{nonparametric regression}, where the regression function $m(x) = \mathbb{E}[Y \mid X=x]$ is estimated without parametric assumptions. One widely used class of estimators is \emph{kernel smoothing}. The Nadaraya--Watson estimator is defined as
\begin{equation}
    \label{eqn:encludea_kernel_smoothing}
    \hat{m}(x) = \frac{\sum_{i=1}^n K_h(x - X_i) Y_i}{\sum_{i=1}^n K_h(x - X_i)},
\end{equation}
where $K_h(u) = h^{-d}K(u/h)$ is a kernel function with bandwidth $h$. Under mild smoothness assumptions, kernel estimators achieve the minimax-optimal convergence rates for nonparametric regression \citep{tsybakov2009introduction}. Beyond kernel methods, local polynomial estimators provide bias reduction while retaining similar variance properties \citep{fan1996local}. These methods form the classical toolkit for nonparametric regression, and they inspire our extension to regression in the space of probability measures (cf. Section~\ref{subsec:local_estimator} below). The analogy is that, just as kernel smoothing locally averages Euclidean responses, one may seek analogous ``local averaging'' or ``local transport'' operations in Wasserstein space to estimate regression functions between distributions.

\section{Methodology}

In this section, we first introduce a nonparametric analog of the standard regression model $Y = f(X) + \varepsilon$ in the Euclidean space where $f : \mathbb{R}^d \to \mathbb{R}$ is the regression function of interest and $\varepsilon$ is a mean-zero noise term. Then, we present our local kernel smoothing estimator based on the Wasserstein geometry. After that, we will describe the neural network architecture that will be used to solve the localized transportation maps for the nonparametric Wasserstein regression.

\subsection{Wasserstein Nonparametric Regression Model}
\label{subsec:regression_model}

Let $\mu \in \mathcal{P}_2(\mathbb{R}^d)$ be a source probability measure (i.e., the reference measure). We define the nonparametric regression model operating between the source and target Wasserstein spaces as
\begin{equation}\label{eq:regressionmodel}
    \nu = T_{\varepsilon}\#(T_{\mu}\#\mu),
\end{equation} 
where $T_{\mu}(x) \coloneqq T(\mu,x) : \mathcal{P}_2\left(\mathbb{R}^d\right)\times\mathbb{R}^d \to \mathbb{R}^d$ is the signal map at $\mu$ and \(T_{\varepsilon}:\mathbb{R}^d \to \mathbb{R}^d\) represents the random noise perturbation satisfying the \emph{unbiasedness} criterion, i.e., the mean-preserving condition
\[
\mathbb{E} \left[T_{\varepsilon}(x)\right] = x, \quad \forall x \in \mathbb{R}^d.
\]
Here, $\#$ is the pushforward operation. In what follows, we focus on estimating the signal map \(T_{\mu}\) at any given source measure $\mu$ from observed data where \(\{\left(\mu_i,\nu_i\right)\}_{i=1}^n\) are independently sampled from model~\eqref{eq:regressionmodel} via $\nu_i = T_{\varepsilon_i}\#(T_{\mu_i}\#{\mu_i})$ with i.i.d. sampled noise maps $T_{\varepsilon_i}$ for $i = 1,\dots,n$.

Our proposed model \eqref{eq:regressionmodel} is different from the regression model considered in the previous works \citep{pmlr-v28-oliva13,ghodrati2022distribution,ghodrati2023transportation}, where the regression operator is defined as
\begin{equation}\label{eqn:global_regressionmodel}
    \nu = T_{\varepsilon}\#(T_0\#\mu),
\end{equation}
for some unknown transport map $T_0 : \mathbb{R}^d \to \mathbb{R}^d$. Note that the signal map $T_0$ in~\eqref{eqn:global_regressionmodel} does not depend on the input source measure $\mu$. Thus, it is a direct analogue of the Euclidean scalar-on-vector regression by viewing the transport map $T_0$ as the ``regression function" and is only capable of capturing the global relationship between $\mu$ and $\nu$.

Clearly, model~\eqref{eqn:global_regressionmodel} is a special case of (and thus less flexible than) our proposed model~\eqref{eq:regressionmodel} since the latter allows the transport map to depend and reflect some features from the source domain. A simple counterexample that model~\eqref{eqn:global_regressionmodel} cannot cover is when both source and target distributions $\mu$ and $\nu$ are discrete over $\mathbb{R}$ supported at two points 0 and 1. Then model~\eqref{eqn:global_regressionmodel} implies that either $\nu(1) = \mu(1)$ or $\nu(1) = \mu(0)$, which does not allow $\nu(1)$ to be as a general function of $\mu(1)$ or $\mu(0)$. In contrast, our model~\eqref{eq:regressionmodel} can be viewed as regressing the optimal transport map $T_{\mu}$ from $\mu$ to $\nu$ (as the response) on $\mu$ as the covariate.

We emphasize that the proposed model~\eqref{eq:regressionmodel} is more flexible than operator learning approaches~\citep{GracykChen2022,10.5555/3618408.3618442}, which aim to learn the mapping from source to target measures. In our setting, although the noise maps $T_{\varepsilon_i}$ are i.i.d., the sampling model~\eqref{eq:regressionmodel} produces data pairs $\{(\mu_i,\nu_i)\}_{i=1}^n$ that are independent but \emph{not identically distributed}. Moreover, the covariate-adjusted interpretation of the local map $T_{\mu}$ naturally entails distribution shift from training data when making predictions on unseen $\mu$, a phenomenon that neural operator models are not equipped to handle.

\subsection{Local Smoothing Estimator}
\label{subsec:local_estimator}

Now, we consider a supervised learning procedure to estimate the transport maps $\{T_{\mu}\}_{\mu \in \mathcal{P}_2(\mathbb{R}^d)}$ from pairs of the observed data \(\{\left(\mu_i,\nu_i\right)\}_{i=1}^n\). Let us fix a reference measure \(\mu\), and our goal is to learn the map \(T_{\mu}\). Our starting point is the well-known variational formulation of the classical kernel smoothing estimator~\eqref{eqn:encludea_kernel_smoothing} as
\begin{equation}
    \label{eqn:eluclidean_variational_form}
    \hat{m}(x) := \arg\min_{c \in \mathbb{R}} \sum_{i=1}^n (c - Y_i)^2 K\Big( {\|x - X_i\| \over h} \Big),
\end{equation}
where $\|\cdot\|$ is the Euclidean norm on $\mathbb{R}^d$. Note that the variational form only requires the input features/predictors to stay in a metric space (i.e., a vector space structure is not needed). Thus,~\eqref{eqn:eluclidean_variational_form} can be naturally extended to a local smoothing estimator under the Wasserstein geometry $(W_2, \mathcal{P}_2(\mathbb{R}^d))$. Specifically, we propose to minimize the following empirical loss function as 
\begin{equation}\label{eq: empirical-loss}
     \hat{T}_{\mu} := \arg\min_{T : \mathbb{R}^d \to \mathbb{R}^d} \left\{ \mathcal{L}_n(T) \coloneqq \frac{1}{n}\sum_{i=1}^n W_2^2\left(T\#\mu_i,\nu_i\right) K_h(\mu,\mu_i) \right\},
\end{equation}
where $K_h(\mu_1, \mu_2) = h^{-d} K(W_2(\mu_1, \mu_2) / h)$ and $K(u)$ is a kernel function satisfying $K(u) \geq 0$ and $\int_{\mathbb{R}} K(u) du= 1$. The kernel function \(K_h(\mu, \mu_i)\) plays a central role in localizing the estimator around a reference measure \(\mu\). It down-weights the influence of training pairs \((\mu_i, \nu_i)\) that are far from \(\mu\) under the 2-Wasserstein distance, ensuring that the estimated map \(\hat{T}_{\mu}\) reflects the local geometry of the regression problem. This design promotes flexibility in learning distribution-to-distribution relationships that may vary across the space of input measures.

Making analogy to the local linear estimators for classical nonparametric regression in the Euclidean space~\citep{fan1996local}, if we parameterize \(T_{\mu}(x) = \alpha(\mu) + B(\mu)x,\) then we obtain a Wasserstein local linear regression estimator via~\eqref{eq: empirical-loss}. On the other hand, since the local linear regression has the curse-of-dimensionality~\citep{tsybakov2009introduction} and our primary goal is to model the transport maps between two infinite-dimensional source and target Wasserstein spaces, we adopt a neural network approach as our functional approximator for the local regression operators (cf. Section~\ref{subsec:neural_network_architecture} for more details).

Once trained, each local estimator \(\hat{T}_{\mu}\) can be reused at test time to make predictions for new source distributions \(\mu'\) that are close to \(\mu\), by applying the learnt map \(\hat{T}_{\mu}\) to samples from \(\mu'\). In practice, we train multiple local models centered at a collection of reference measures \(\{\mu_0^{(l)}\}_{l=1}^M\), and for a given test distribution, predictions can be made using the nearest \(\mu_0^{(l)}\) or a weighted combination of nearby local estimators. This localized approach allows us to approximate complex global regression structures using a set of simpler, interpretable affine transformations.

\subsection{Kernel Bandwidth Tuning}\label{sec: bw}

The hyperparameter, specific to our framework, that required tuning, aside from standard choices such as learning rate or batch size, was the kernel bandwidth $h$. 
Across all experiments (Gaussian, Gaussian mixture, and MNIST in Section~\ref{sec: simstudies}), we employed a nearest-neighbor heuristic to select $h$ in a data-adaptive manner. 
Specifically, let $\{\mu_i\}_{i=1}^n$ denote the training source distributions, and fix a test distribution $\mu_0$, and compute pairwise distances.
For a set number of neighbors $k \ll n$, let 
\[
r_k(\mu_0) \;=\; \min\left\{ r : \sum_{i=1}^n\mathbf{1}\left\{ i : d(\mu_0,\mu_i) \le r \right\} \ge k \right\},
\]
the distance to the $k$-th nearest neighbor of $\mu_0$. 
We then set the kernel bandwidth
\[
h(\mu_0) \;=\; \rho \, r_k(\mu_0),
\]
with scaling constant $\rho>0.$ Typically $\rho=1,$ but is adapted as the dimension increases to account for the curse of dimensionality.

In words, the bandwidth is chosen so that the effective kernel support includes approximately $k$ training distributions around $\mu_0$. 
This ensures that the regression estimator is smoothed appropriately while adapting to the local density of training distributions in Wasserstein space, even in settings when the data is highly irregular. The same principle was used across all simulation settings with slight modifications to adjust to differences in Gaussian, Gaussian Mixture, and MNIST data.

\subsection{Neural Network Architectures}
\label{subsec:neural_network_architecture}

In this section, we describe how these estimators are learned in practice using a neural networks tailored to each setting. We utilize two different neural net architectures, depending on the task. For source and target pairs of empirical distributions, we employ a DeepSets neural network architecture \citet{zaheer2018deepsets} to handle the permutation invariant nature of the given data. Moreover, we employ U-Nets \citet{ronneberger2015u} as the task becomes an image-to-image regression learning task. 

\subsubsection{DeepSets}\label{section: deepsets}

To estimate local transport maps, we adopt a nonparametric, local regression framework in which a separate neural network is trained for each chosen reference distribution \(\mu_0^{(l)}\), using only training pairs that are close to \(\mu_0^{(l)}\) in the Wasserstein sense. Each network learns to output a transport map \(T:\mathbb{R}^d \to \mathbb{R}^d\) in one of two variants:
\[
T(x) = \alpha + Bx, \qquad
T(x) = x + \Delta(x,z),
\]
where \((\alpha,B)\) parameterize an affine transformation and \(\Delta(x,z)\) denotes a more general local map, where the network conditions on the distribution via a global context vector \(z\), and predicts per-point displacements.  
The learning objective is a Sinkhorn-approximated Wasserstein-2 loss, weighted by the kernel \(K_h(\mu_0^{(l)}, \mu_i)\).

Let \(\mu_i\) and \(\nu_i\) denote source and target distributions, respectively, and suppose we observe \(k\) i.i.d.\ samples from each,
\[
x^{(i)}_j \overset{\text{iid}}{\sim} \mu_i, 
\qquad 
y^{(i)}_j \overset{\text{iid}}{\sim} \nu_i, 
\qquad j = 1,\ldots,k.
\]
We define empirical measures
\[
\hat{\mu}_i = \frac{1}{k} \sum_{j=1}^k \delta_{x^{(i)}_j}, 
\qquad 
\hat{\nu}_i = \frac{1}{k} \sum_{j=1}^k \delta_{y^{(i)}_j}.
\]

Each network uses a DeepSets-style architecture~\citep{zaheer2018deepsets} to process the input distribution \(\hat{\mu}_i\) by its point cloud representation \(X_i = \{x^{(i)}_1, \ldots, x^{(i)}_k\} \subset \mathbb{R}^d\). The encoder–decoder structure is
\[
f_l(X_i) = \rho_l\!\left( \frac{1}{k} \sum_{j=1}^k \psi_l(x^{(i)}_j) \right),
\]
where \(\psi_l: \mathbb{R}^d \to \mathbb{R}^h\) is a pointwise encoder (two-layer MLP with ReLU), the sum produces a permutation-invariant aggregation, and \(\rho_l\) is a decoder whose output depends on the chosen map variant.

In the affine case, the decoder outputs
\[
(\alpha_i^{(l)}, B_i^{(l)}) = \rho_l(z_i),
\quad 
\alpha_i^{(l)} \in \mathbb{R}^d, 
\quad 
B_i^{(l)} \in \mathbb{R}^{d\times d}.
\]
In the general local map case, the decoder produces a global context vector \(z_i\), which is concatenated with each point \(x\) and passed through an additional \textit{displacement head} to compute per-point corrections \(\Delta(x,z_i)\).

Thus, for each reference distribution \(\mu_0^{(l)}\), the network learns a local model \(\hat{T}_{\mu_0^{(l)}}\) such that \(\hat{T}_{\mu_0^{(l)}}\# \hat{\mu}_i \approx \hat{\nu}_i\). This procedure is identical whether the distributions are Gaussians, Gaussian mixtures, or empirical distributions from real data, since in all cases the model only sees their sampled point clouds and the architecture is inherently permutation-invariant.

\subsubsection{U-Net}\label{sec:unet}

For distributions supported on regular image grids (e.g.\ MNIST digits), we employ a compact U\!-Net architecture~\citep{ronneberger2015u}, which is natural for image-to-image regression tasks. 
Unlike the permutation-invariant DeepSets model, the U\!-Net exploits the underlying spatial structure of the data through convolutions and skip connections.
Let $x \in \mathbb{R}^{1 \times H \times W}$ denote a grayscale input image. 
The network $f_\theta : \mathbb{R}^{1 \times H \times W} \to \mathbb{R}^{1 \times H \times W}$ follows the schematic form
\[
x \;\xrightarrow{E_1}\; s_1 \;\xrightarrow{E_2}\; s_2 \;\xrightarrow{M}\; z 
\;\xrightarrow{\text{upsample}\times 2}\; \tilde{z} 
\;\xrightarrow{[\tilde{z};\, s_1]}\; D_1 
\;\xrightarrow{\text{output}}\; f_\theta(x),
\]
where $[\cdot;\cdot]$ denotes channel-wise concatenation.

$E_1$ and $E_2$ are the encoder blocks, progressively reducing spatial resolution while increasing the number of feature channels, $M$ is the bottleneck at the coarsest resolution, the upsampling stage restores spatial resolution, and $D_1$ is the decoder block that fuses the upsampled features $\tilde{z}$ with the high-resolution skip connection $s_1$. 
The output head then maps the decoder features to a prediction $f_\theta(x)$.

This U\!-Net is lightweight, with a single downsampling stage, one bottleneck, and a symmetric upsampling path. 
Skip connections link encoder and decoder features at matching resolutions, enabling the model to combine local detail ($s_1$) with global context ($z$). 
Convolutions are followed by normalization and nonlinear activations, and residual connections are used within blocks to stabilize training. 
The final layer outputs an image of the same resolution as the input, which we interpret as a probability distribution after applying a softplus transform and per-image normalization. 

\section{Simulation Studies}\label{sec: simstudies}

We conducted experiments using three types of datasets, Gaussian, Gaussian mixture, and MNIST image pairs. Model hyperparameters, include learning rate, batch size, number of epochs, kernel bandwidth $h$ (discussed in Section~\ref{sec: bw}), and the blur parameter $\varepsilon$ associated with the Sinkhorn approximation used to estimate the Wasserstein distance. All models were trained using the Adam optimizer with default momentum parameters, a learning rate of $10^{-3}$, and the blur parameter was fixed $\varepsilon=0.15$, unless otherwise noted. Moreover, evaluation was performed on unseen test pairs.

\subsection{Gaussian Experiments}
\begin{figure}[!htbp]
    \centering
    \begin{subfigure}[t]{0.3\linewidth}
        \centering
        \includegraphics[width=\linewidth]{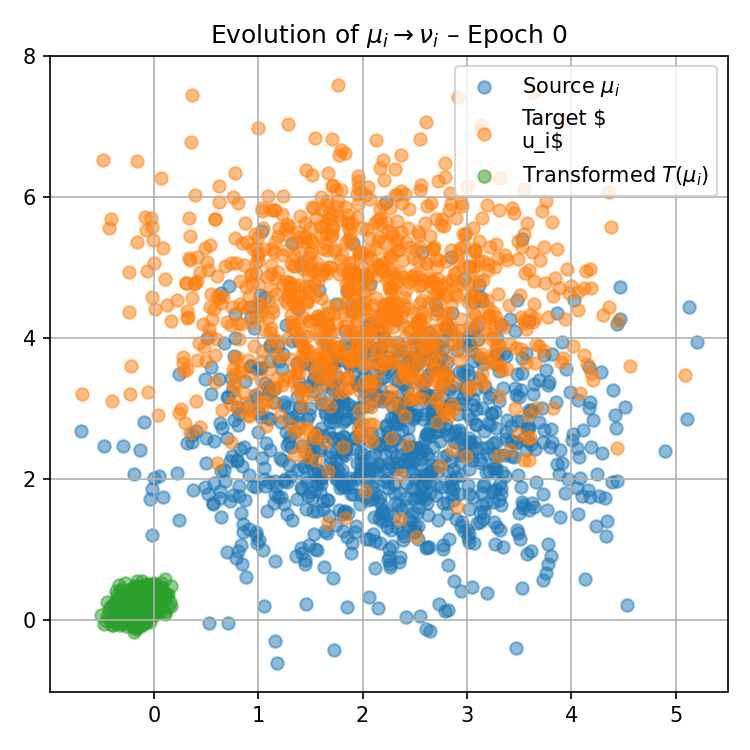}
        \caption{}
        \label{fig:pushforward-epoch0}
    \end{subfigure}
    \hfill
    \begin{subfigure}[t]{0.3\linewidth}
        \centering
        \includegraphics[width=\linewidth]{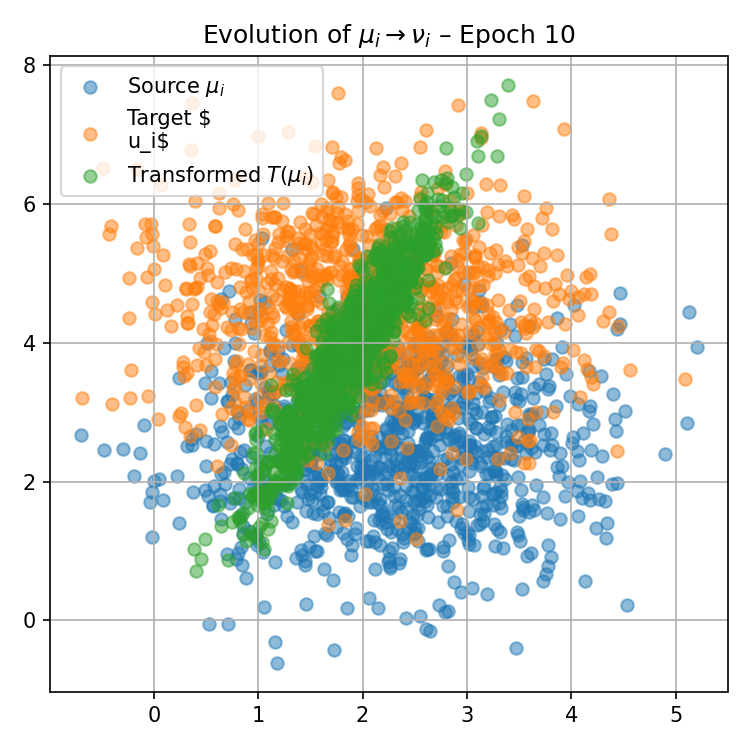}
        \caption{}
        \label{fig:pushforward-epoch10}
    \end{subfigure}
    \hfill
    \begin{subfigure}[t]{0.3\linewidth}
        \centering
        \includegraphics[width=\linewidth]{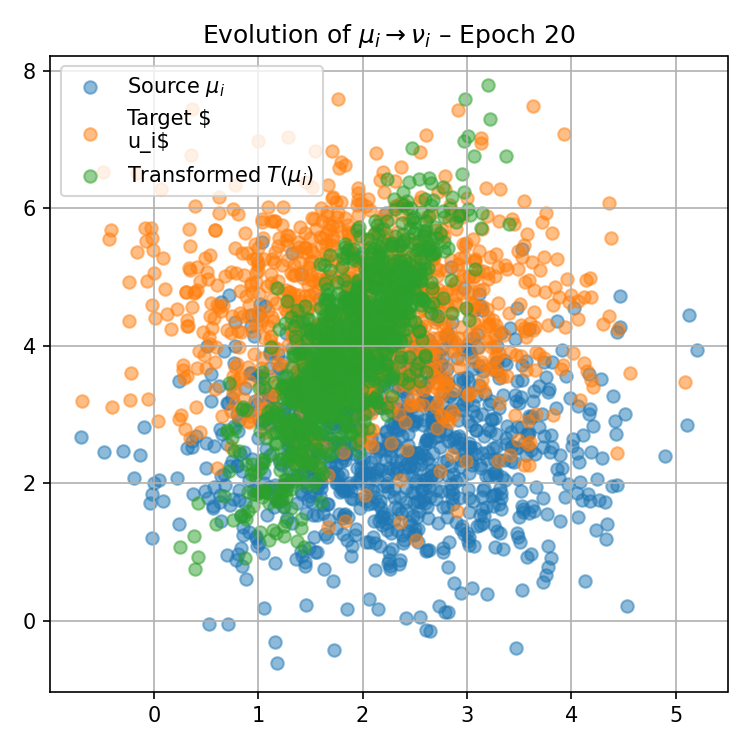}
        \caption{}
        \label{fig:pushforward-epoch20}
    \end{subfigure}

    \vspace{0.5em}  

    \begin{subfigure}[t]{0.3\linewidth}
        \centering
        \includegraphics[width=\linewidth]{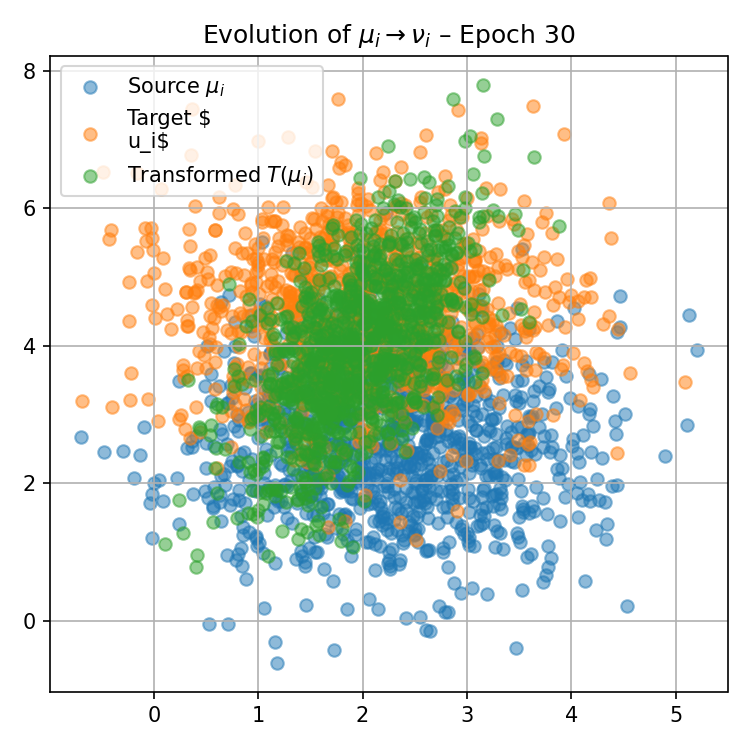}
        \caption{}
        \label{fig:pushforward-epoch30}
    \end{subfigure}
    \hfill
    \begin{subfigure}[t]{0.3\linewidth}
        \centering
        \includegraphics[width=\linewidth]{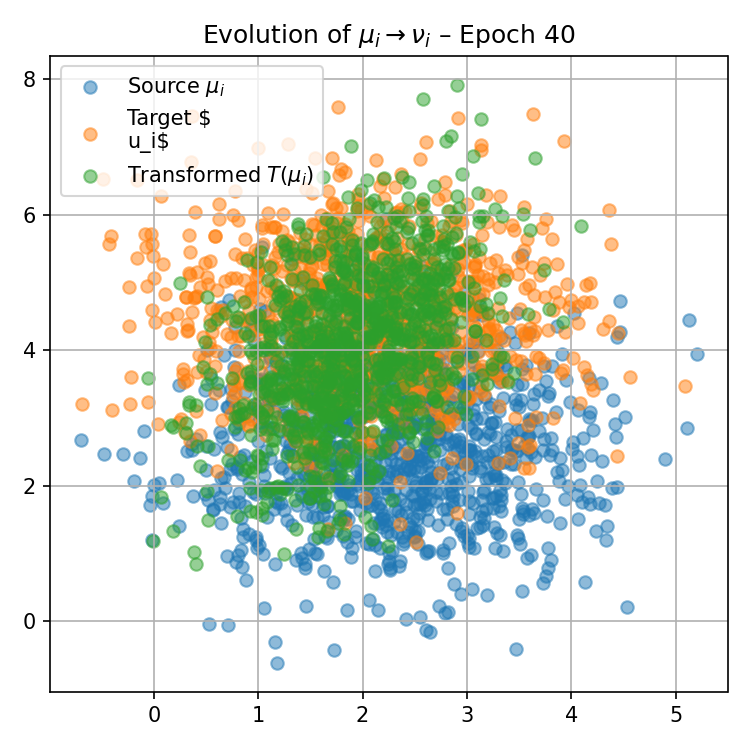}
        \caption{}
        \label{fig:pushforward-epoch40}
    \end{subfigure}
    \hfill
    \begin{subfigure}[t]{0.3\linewidth}
        \centering
        \includegraphics[width=\linewidth]{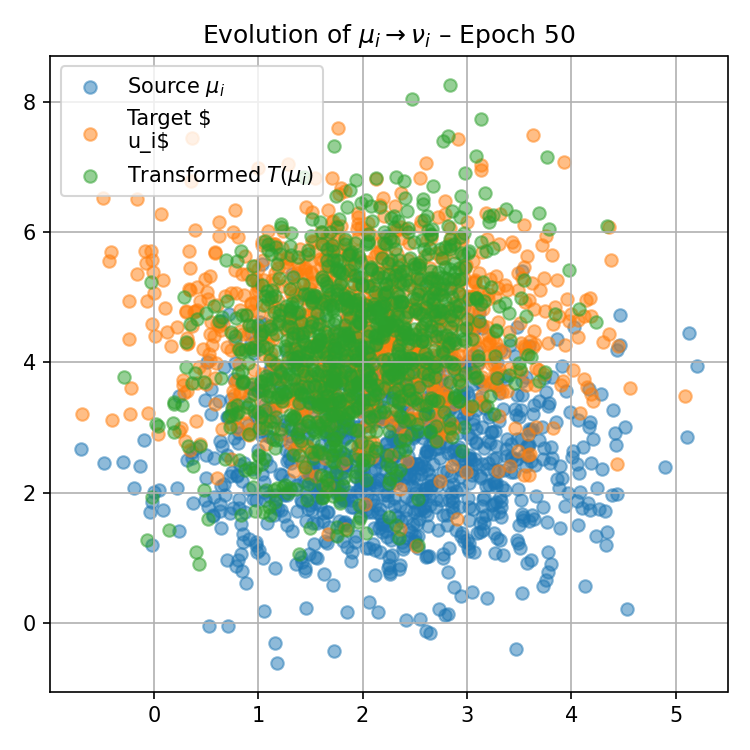}
        \caption{}
        \label{fig:pushforward-epoch50}
    \end{subfigure}

    \caption{Progression of pushforward distributions for a fixed reference measure $\mu_0^{(0)}$ over training epochs.}
    \label{fig:pushforward-grid}
\end{figure}

We trained a family of local models to pushforward source distributions \(\mu_i\) toward their targets \(\nu_i\), using a kernel-weighted Sinkhorn loss centered at several reference distributions \(\mu_0^{(l)}\). Details for the data generation are available in Sec~\ref{sec: gauss-data}. Each local model was trained independently using only those training pairs \((\mu_i, \nu_i)\) that were close, in the 2-Wasserstein sense, to a given reference \(\mu_0^{(l)}\). Note, in later experiments, we change the strategy slightly.  

In these early experiments, when evaluating the performance of our methods on simple Gaussian distributions, we found that even simple affine maps \(T(x) = Bx + \alpha\) were capable of producing good visual alignment between the pushed-forward source distribution and the target as seen in Figure [\ref{fig:pushforward-grid}]. We trained the neural network introduced in Section~\ref{section: deepsets} to learn the parameters \(B\) and \(\alpha\).

\subsection{Gaussian Mixtures Experiments}

In the Gaussian mixture experiments, the estimator is constructed in a local fashion. 
Given a fixed test source distribution $\mu_0$ with corresponding target $\nu_0$, and a collection of training pairs $\{(\mu_i,\nu_i)\}_{i=1}^n$, the goal is to learn a local transport map $\hat{T}_{\mu_0}$ such that
\[
\hat{T}_{\mu_0} \# \mu_0 \;\approx\; \nu_0.
\]
The estimator is trained using only those training pairs $(\mu_i,\nu_i)$ that lie within the effective support of the kernel centered at $\mu_0$, with weights $K_h(\mu_0,\mu_i)$. For this setting, we use the generalized DeepSets architecture from Section~\ref{section: deepsets} that outputs an empirical prediction $\hat{\nu}_0$ directly from the point cloud representation of $\mu_0$ and its kernel-weighted neighbors.

The description for the generated data used in this setting is available in Section~\ref{sec: gmm-data}. We store the training data in tensors
\[
\texttt{mu} \in \mathbb{R}^{N\times K\times 2}, \qquad
\texttt{nu} \in \mathbb{R}^{N\times K\times 2},
\]
as a \emph{master dataset} (e.g., $N=1000$, $K=10{,}000$).
Training regimes $(n,k)$ are created by deterministic subsampling using a subset seed. Each Regime is run $10$ times by subsampling the master dataset with a different seed. 

\begin{wraptable}{r}{0.6\textwidth}
\vspace{-0.75em}
\centering
\caption{Average test error and Wasserstein coefficient of determination $R_W^2$ (Eq.~\eqref{eq:R2W_def}) across simulation regimes.}
\label{tab:regime_results}
\scriptsize
\setlength{\tabcolsep}{6pt}
\renewcommand{\arraystretch}{1.1}
\begin{tabular}{@{}rrrrc@{}}
\toprule
$d$ & $n$ & $k$ & Absolute Test Error & $R_W^2$ \\
\midrule
2 & 100  & 1000 & $0.01922 \,\pm\, 0.00694$ & $0.99977$ \\
2 & 100  & 100  & $0.02438 \,\pm\, 0.00995$ & $0.99971$ \\
2 & 10   & 1000 & $0.68470 \,\pm\, 0.34196$ & $0.99190$ \\
2 & 1000 & 1000 & $0.00124 \,\pm\, 0.00011$ & $0.99983$ \\
5 & 100  & 1000 & $2.34761 \,\pm\, 0.03216$ & $0.94945$ \\
5 & 100  & 100  & $2.36256 \,\pm\, 0.03350$ & $0.94913$ \\
5 & 100  & 10   & $23.07784 \,\pm\, 1.30152$ & $0.94064$ \\
5 & 10   & 100  & $19.52894 \,\pm\, 3.72827$ & $0.95291$ \\
\bottomrule
\end{tabular}
\vspace{-0.5em}
\end{wraptable}
\par

Let \(\{(\mu_i, \nu_i)\}_{i=1}^n\) be training pairs, and let \(\hat{T}_{\mu_0}\) denote the estimated local transport map used to predict \(\nu_0\) from \(\mu_0\).
Define the empirical Wasserstein barycenter of the target measures as
\[
\bar{\nu}_W = \arg\min_{\nu \in \mathcal{P}_2(\mathbb{R}^d)} \sum_{i=1}^n W_2^2(\nu, \nu_i).
\]
Then, we define
\(
SS_{\text{res}} = W_2^2\!\big(\nu_0,\, \hat{T}_{\mu_0}\#\mu_0\big),
\qquad
SS_{\text{tot}} = W_2^2\!\big(\nu_0,\, \bar{\nu}_W\big),
\)
and the \emph{Wasserstein coefficient of determination} as
\begin{equation}
R_W^2 = 1 - \frac{SS_{\text{res}}}{SS_{\text{tot}}}
      = 1 - \frac{W_2^2(\nu_0,\, \hat{T}_{\mu_0}\#\mu_0)}
                     {W_2^2(\nu_0,\, \bar{\nu}_W)}.
\label{eq:R2W_def}
\end{equation}

In the two-dimensional setting, our estimator achieves near-exact recovery provided sufficient sample size (\(n \geq 100\)), with relative errors on the order of \(10^{-4}\). 
Performance deteriorates rapidly with small sample size (\(n=10\)), exacerbated by insufficient kernel support.
In higher dimension (\(d=5\)), both absolute and relative errors increase by one to two orders of magnitude, with relative errors saturating around \(0.65\)--\(0.70\). To keep results consistent, the bandwidth used in the two-dimensional cases was recycled in the five-dimensional cases, only scaled by a factor to adjust for the increase in dimension. Thus, it is important to note that the training could have potentially been optimized for the five-dimensional case. The results indicate possible sensitivity to the curse of dimensionality and highlight the importance of appropriate bandwidth scaling. 
Overall, these results demonstrate that the method is highly effective in low-dimensional regimes with adequate data, but requires careful tuning and larger sample sizes to remain stable as dimension increases.

\subsection{MNIST Experiment}

\begin{wrapfigure}[28]{r}{0.55\textwidth}  
    \centering
    \vspace{-1em}
    \setlength{\tabcolsep}{2pt}
    \renewcommand{\arraystretch}{0.55}
    {\scriptsize
    \begin{tabular}{ccc}
        \includegraphics[width=0.2\linewidth]{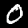} &
        \includegraphics[width=0.2\linewidth]{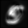} &
        \includegraphics[width=0.2\linewidth]{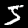} \\
        (a) Input (0) & (b) Prediction (5) & (c) Ground truth (5) \\

        \includegraphics[width=0.2\linewidth]{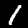} &
        \includegraphics[width=0.2\linewidth]{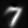} &
        \includegraphics[width=0.2\linewidth]{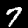} \\
        (d) Input (1) & (e) Prediction (7) & (f) Ground truth (7) \\

        \includegraphics[width=0.2\linewidth]{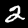} &
        \includegraphics[width=0.2\linewidth]{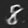} &
        \includegraphics[width=0.2\linewidth]{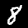} \\
        (g) Input (2) & (h) Prediction (8) & (i) Ground truth (8) \\

        \includegraphics[width=0.2\linewidth]{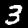} &
        \includegraphics[width=0.2\linewidth]{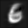} &
        \includegraphics[width=0.2\linewidth]{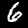} \\
        (j) Input (3) & (k) Prediction (6) & (l) Ground truth (6) \\

        \includegraphics[width=0.2\linewidth]{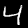} &
        \includegraphics[width=0.2\linewidth]{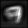} &
        \includegraphics[width=0.2\linewidth]{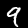} \\
        (m) Input (4) & (n) Prediction (9) & (o) Ground truth (9) \\
    \end{tabular}
    } 
    \caption{\scriptsize MNIST transformation results across multiple digit pairs. Each row shows a source digit (left), predicted output (middle), and ground-truth target (right). Images are treated as normalized discrete probability distributions while evaluating loss.}
    \label{fig:mnist-results}
    \vspace{-1em}
\end{wrapfigure}
\par

In this experiment, we evaluate our method on the benchmark MNIST dataset of handwritten digits, constructing a subset designed to test distributional regression in a heterogeneous setting. Specifically, we select five digit pairs: 
\((0 \to 5), (1 \to 7), (2 \to 8), (3 \to 6), (4 \to 9)\). For each pair, all images of the source digit serve as source distributions $\{\mu_i\}_{i=1}^n$, while the corresponding images of the target digit serve as target distributions $\{\nu_i\}_{i=1}^n$. 
From each pair we reserve one source–target example $(\mu_0^a, \nu_0^b)$ for testing, and use the remainder for training. 
The U\!-Net described in Section~\ref{sec:unet} is trained to learn local maps $T_{\mu_0^a}$, which are then evaluated by predicting
\[
\nu_0^b \;\approx\; T_{\mu_0^a}\#\mu_0^a,
\]
for each reserved test pair.

In all experiments we used $N_{\text{per class}}=1000$ samples.  Each $28\times 28$ grayscale image was mapped to the unit square $[0,1]^2$ by assigning normalized pixel intensities to the centers of the pixels, thereby representing each image as a discrete probability distribution. 
The Wasserstein kernel loss was then computed directly on these empirical measures. We made a strong assumption that all images of one type would be closer to each other in the Wasserstein sense despite a large variation in how images are handwritten. This proved to hold in all but one case, and, in that case, we trimmed some percentages off entirely to concentrate the signal.

\section{Discussion and Limitations}
\label{sec:discussion}

This work introduces \emph{Neural Local Wasserstein Regression}, a nonparametric DoD regression framework that models transport via locally defined maps, trained with kernel weights in $W_2$ space and instantiated with permutation-invariant (DeepSets) and convolutional (U\!-Net) architectures. The results suggest that local transport improves approximation capacity relative to global map or linearization-based approaches when the predictor space exhibits heterogeneous geometry. Practically, local training reduces the reliance on a single global map that may not exist or be stable, enables reuse of fitted maps in the vicinity of a reference measure, and supports modular deployment (multiple local models can be trained and composed or selected at test time). 

Local estimators rely on the kernel bandwidth $h$ and the effective number of neighbors. While our $k$NN heuristic adapts to local density, it remains a tuning knob. Key questions about identifiability of $T_\mu$, consistency and rates under local smoothness of the transport field, and the effect of $h$ on statistical and computational error will be pursued in future work.


\appendix

\section{Technical Appendices and Supplementary Material}

\subsection{Data Generation}
\subsubsection{Gaussian Synthetic Data Generation}\label{sec: gauss-data}
To evaluate the performance of our distribution-to-distribution regression framework, we first construct a synthetic Gaussian dataset in \(\mathbb{R}^2\) with ground truth maps that are smoothly parameterized.

For each distribution pair \((\mu_i, \nu_i)\), we generate data as follows:

\begin{enumerate}
    \item Sample a mean vector \(m_i \sim \text{Uniform}([-3, 3]^2)\).
    \item Generate source samples \(x^{(i)}_j \overset{\text{iid}}{\sim} \mathcal{N}(m_i, I_2)\), and define \(\mu_i = \frac{1}{n} \sum_{j=1}^n \delta_{x^{(i)}_j}\).
    \item Define a rotation angle \(\theta_i = 2 \|m_i\|\), and construct the corresponding rotation matrix:
    \[
    A_i = \begin{bmatrix}
        \cos(\theta_i) & -\sin(\theta_i) \\\\
        \sin(\theta_i) & \cos(\theta_i)
    \end{bmatrix}.
    \]
    \item Define a translation vector \(a_i = 0.5 \cdot m_i\).
    \item Define the ground truth affine map \(T_i(x) = A_i x + a_i\).
    \item Apply the map to each source sample and add Gaussian noise \(\varepsilon_i \sim \mathcal{N}(0, \sigma^2 I_2)\) to produce the target samples:
    \[
    y^{(i)}_j = T_i(x^{(i)}_j) + \varepsilon_i, \quad \nu_i = \frac{1}{n} \sum_{j=1}^n \delta_{y^{(i)}_j}.
    \]
\end{enumerate}

The resulting dataset \(\{(\mu_i, \nu_i)\}_{i=1}^n\) consists of empirical distributions related by smooth, locally varying affine maps. This setup reflects a structured but nontrivial regression problem while ensuring that the true map varies smoothly as a function of the source distribution.

\subsubsection{Synthetic Gaussian Mixture Data Generation}\label{sec: gmm-data}

We generate training pairs $\{(\mu_i,\nu_i)\}_{i=1}^n \subset \mathcal{P}_2(\mathbb{R}^2)$ where each $\mu_i$ is an empirical Gaussian mixture and $\nu_i$ is obtained by a piecewise smooth map applied to $\mu_i$.

Fix $C \in \mathbb{Z}^+$ to be the number of mixture components, and $k \in \mathbb{Z}^+$ to be the number of samples per distribution. To generate the source distributions, $\mu_i,$ for each $i \in \{1,\dots,n\},$ we begin by sampling the mixture weights $w_i \sim \mathrm{Dirichlet}(\mathbf{1}_C).$ Then, we sample component means $m_{ij} \sim \mathrm{Unif}([L,U]^2))$ for $j=1,\dots,C,$ and draw component indices $c_\ell \sim \mathrm{Categorical}(w_i)$ for $\ell=1,\dots,k,$ sample points with isotropic covariance $\sigma^2 I_2$:
\[
x_{i\ell} \sim \mathcal{N}\!\big(m_{i,c_\ell}, \ \sigma^2 I_2\big), \qquad \ell=1,\dots,k.
\]
The empirical source distribution is
\[
\mu_i = \frac{1}{k}\sum_{\ell=1}^k \delta_{x_{i\ell}}.
\]
For each $\mu_i$, define the barycenter $\bar m_i = \tfrac{1}{k}\sum_{\ell} x_{i\ell}$, and the RMS radius (equivalent to $W_2(\mu_i,\delta_0)$ in the continuum limit)
\[
r_i \coloneqq \Big(\tfrac{1}{k}\sum_{\ell=1}^k \|x_{i\ell}\|^2\Big)^{1/2}.
\]

Rotation angle $\theta(r) = \alpha r$, shear strength $k(r) = \kappa_0 + \kappa_1 r$.  
The matrices
\[
R(\theta) =
\begin{bmatrix}
\cos\theta & -\sin\theta \\
\sin\theta & \cos\theta
\end{bmatrix}, \qquad
S(k) =
\begin{bmatrix}
1 & k \\
0 & 1
\end{bmatrix}
\]
represent rotation and shear, respectively. Then, we define the operator,
\[
A(r) = \lambda(r)\, R(\theta(r)) + (1-\lambda(r))\, S(k(r)),
\]
such that $\lambda(r)\in(0,1)$ centered at $r_{\mathrm{thresh}}$ and 
\[
\lambda(r) = \sigma\!\left(\tfrac{r_{\mathrm{thresh}} - r}{\gamma}\right),
\]
where $\gamma > 0$ controls the transition width. Then, with $a_i = \beta\,\bar m_i$, the samples are mapped by
\[
y_{i\ell} = A(r_i)\, x_{i\ell} + a_i + \varepsilon_{i\ell}, \qquad 
\varepsilon_{i\ell} \sim \mathcal{N}(0,\ \tau^2 I_2),
\]
forming the empirical target distribution
\[
\nu_i = \frac{1}{K}\sum_{\ell=1}^K \delta_{y_{i\ell}}.
\]

\bibliographystyle{plainnat}
\bibliography{xhchen_ref}

\end{document}